\pdfoutput=1

\documentclass[11pt]{article}
\usepackage{placeins}
\usepackage[final]{acl}
\usepackage{tikz}
\usepackage{times}
\usepackage{multirow}
\usepackage{longtable}
\usepackage{adjustbox}
\usepackage{makecell}
\usepackage[edges]{forest}
\usepackage{latexsym}
\usepackage[normalem]{ulem}
\usepackage{anyfontsize}
\usepackage{pifont} 

\usepackage[T1]{fontenc}
\usetikzlibrary{arrows.meta}

\usepackage[utf8]{inputenc}

\usepackage{microtype}

\usepackage{inconsolata}

\usepackage{graphicx}

\usepackage{amsmath}
\usepackage{hyperref}

\usepackage{colortbl}
\usepackage{rotating}
\usepackage{array}
\usepackage{color}
\usepackage{rotating}
\usepackage{tabularray}
\usepackage{longtable}
\usepackage{graphicx}
\usepackage{tabularx}
\usepackage[multiple]{footmisc}
\usepackage{bigfoot}

\title{Building Safe GenAI Applications: An End-to-End Overview of Red Teaming for Large Language Models}
\author{
  Alberto Purpura$^{*}$, Sahil Wadhwa$^{*}$, Jesse Zymet\thanks{Equal contribution}, Akshay Gupta, Andy Luo, Melissa Kazemi Rad, \\ \textbf{Swapnil Shinde, Mohammad Shahed Sorower} \\
  Capital One, AI Foundations \\
  \texttt{alberto.purpura@capitalone.com, sahil.wadhwa@capitalone.com, jesse.zymet@capitalone.com}\\ \texttt{akshay.gupta3@capitalone.com, andy.luo@capitalone.com, melissa.kazemirad@capitalone.com} \\ \texttt{swapnil.shinde2@capitalone.com, mohammad.sorower@capitalone.com}}

\begin{document}

\maketitle

\begin{abstract}
    The rapid growth of Large Language Models (LLMs) presents significant privacy, security, and ethical concerns. While much research has proposed methods for defending LLM systems against misuse by malicious actors, researchers have recently complemented these efforts with an offensive approach that involves \textit{red teaming}, i.e., proactively attacking LLMs with the purpose of identifying their vulnerabilities. This paper provides a concise and practical overview of the LLM red teaming literature, structured so as to describe a multi-component system end-to-end. To motivate red teaming we survey the initial safety needs of some high-profile LLMs, and then dive into the different components of a red teaming system as well as software packages for implementing them. We cover various attack methods, strategies for attack-success evaluation, metrics for assessing experiment outcomes, as well as a host of other considerations. Our survey will be useful for any reader who wants to rapidly obtain a grasp of the major red teaming concepts for their own use in practical applications.

\end{abstract}

\section{Introduction}

The popularity and widespread adoption of Large Language Models (LLMs) has been transformative across many industries, ushering in new possibilities for enhancing productivity, decision-making, and user engagement. LLMs are contributing significantly to fields such as finance, healthcare, and legal services where they are being leveraged for tasks such as customer servicing support, clinical notes and contract analysis.
However, the increasing reliance on LLMs brings with it a critical and challenging ethical-moral responsibility: ensuring that the deployed system responds to any possible input in safe or otherwise desirable ways. While LLMs offer remarkable capabilities, they are also vulnerable to various forms of misuse. 
Such attacks could provoke LLMs to generate misinformative, biased, or toxic content \cite{abid2021persistentantimuslimbiaslarge, lin2022truthfulqameasuringmodelsmimic} or expose private information \cite{carlini2021extractingtrainingdatalarge}. Microsoft's Tay, in a high-profile case, was successfully provoked by attackers to send racist or sexually-charged tweets to a large audience \cite{lee2016}.  
A great deal of research on improving LLM safety has been conducted from a defensive standpoint, with investigators developing methods for guardrailing LLMs against potential attacks \cite{dong2024safeguardinglargelanguagemodels}. These attacks, however, must be identified beforehand, which has proven to be challenging -- e.g., GPT-4 was vulnerable to attacks absent from its safety training that were written in low-resource languages \cite{yong2024lowresourcelanguagesjailbreakgpt4}. Investigators have hence turned to complementing defensive efforts with an offensive approach to LLM safety, proposing strategies for \textit{red teaming} LLMs, i.e., proactively attacking or testing LLMs with the purpose of identifying their vulnerabilities. Red teaming is useful for any organization that aims not only to productionize some LLM-supported system, but to effectively anticipate threats to their system and safeguard against them before production.

\begin{figure}[h!]
    \centering 
    \includegraphics[width=1.5\columnwidth]{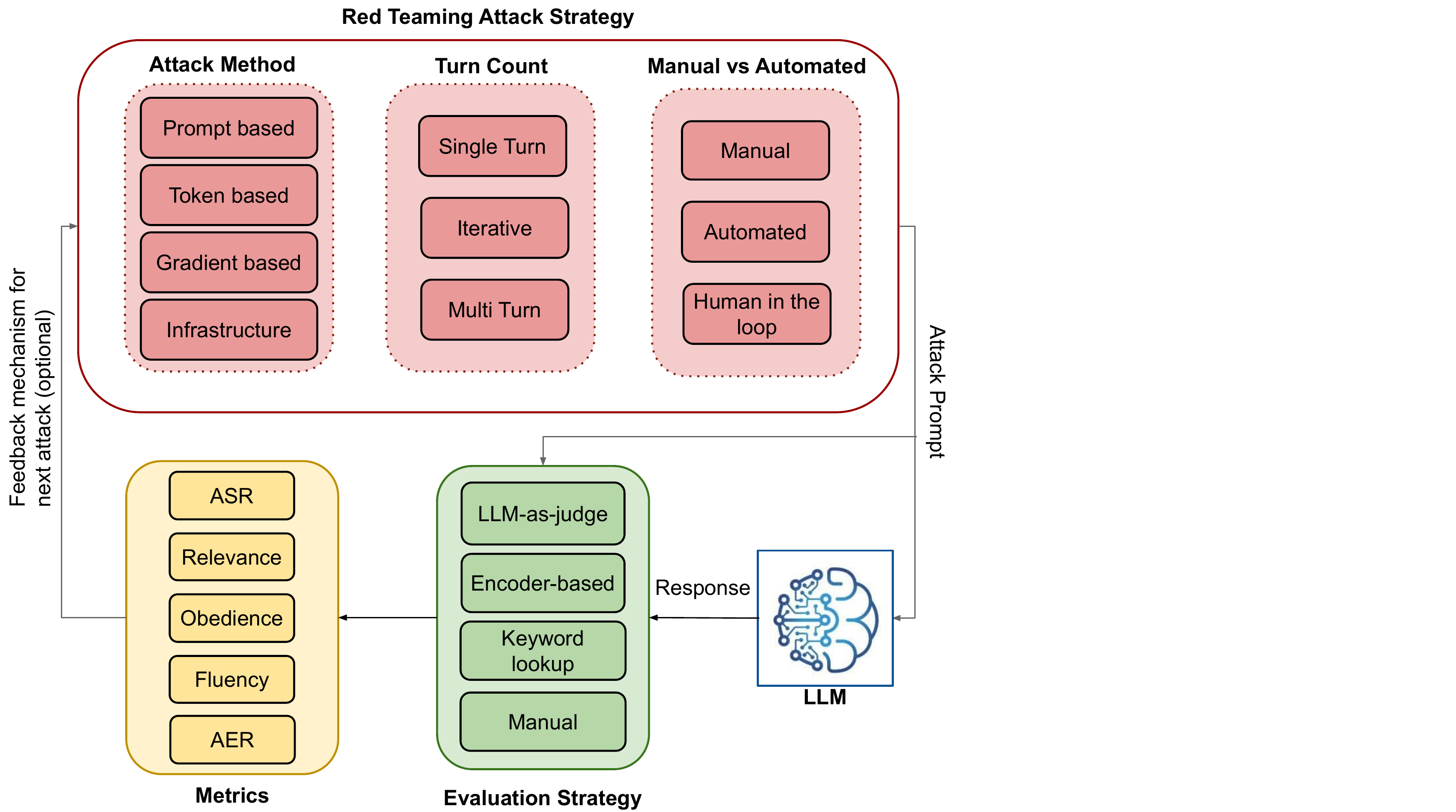} 
    \caption{
        Core components of a red teaming system.
    }
    \label{fig:red_teaming_components}
\end{figure}

While prior reviews of LLM red teaming focused on serving as an encyclopedic taxonomic resource, e.g., of attack methodologies \cite{lin2024against}, we anticipate a wide need for a concise and practical overview geared toward readers who want to rapidly grasp the major concepts and components of a red teaming system and available software tools that have emerged, for example to devise and implement a system of their own. The purpose of this paper is to provide such an overview: one that balances comprehensive treatment of research with conciseness, and structures the exposition to describe a multi-component system end-to-end. Figure \ref{fig:red_teaming_components} provides an illustration of the framework and its components, the latter of which are covered in Sections \ref{sec:categorizing_attacks}, \ref{sec:eval_strategies} and \ref{sec:evaluation_metrics}.
After covering related work, we survey a few case studies from the tech industry, highlighting problems that motivate the field of red teaming. We then dive into the central components that make up a complete red teaming system, reviewing popular methods, software packages, and other resources that have emerged to support these components. We cover various attack strategies, with attention paid to categorizing particular methods and distinguishing single-turn from multi-turn attacking, manual from automated attacking, and different varieties of automated attacking. We then dive into popular approaches to attack success evaluation, as well as safety metrics for assessing overall experimental outcomes. We discuss a number of publicly available resources for red teaming, including software packages and datasets. We also touch briefly upon guardrailing steps commonly taken after red teaming. Finally, we close with future directions that we judge to constitute some of the most impactful opportunities for progress, including strategies for adapting automated attackers to generate more relevant and diverse sets of attacks, including multi-turn ones.

\section{Related work}
Recent literature has explored various facets of LLM red teaming, offering valuable insights into the rapidly evolving field. Some organizations aim to provide up-to-date informational materials geared toward helping developers and web-security practitioners secure their particular applications \cite{mitre_atlas, mlcommons_ai_safety_benchmarks, mlcommons_aisafety_v0_5}. For example, the OWASP Foundation published the OWASP Top 10 \cite{OWASPTop10}, a document that describes, as deemed by common consensus, some of the most major threats to the security of LLM-supported applications, and provides mitigation strategies. On the academic side, \citet{feffer2024red} provide a high-level overview of and stance on red teaming practices, indexing on particular aspects of the literature to argue that the red teaming community lacks consensus around scope, structure, and evaluation of red-teaming. \citet{verma2024operationalizing} operationalize a threat model for red teaming, providing a taxonomy based on entry points in the LLM lifecycle. \citet{rawat2024attack} provide a practitioner's viewpoint of challenges within LLM red teaming and emphasizes the context-dependent nature of vulnerabilities, and introduces a taxonomy of single-turn, prompt-based attacks. \citet{mo2024tremblinghousecardsmapping} develop a taxonomy of attacks against language agents in particular -- i.e., systems equipped with additional capacities for reasoning, planning, and task completion. \citet{shi2024large} offer a comprehensive survey of LLM safety more broadly, encompassing various risks beyond attacks, including value misalignment and autonomous AI risks. But perhaps the most extensive treatment to date specifically of LLM red teaming is given in \citet{lin2024against}, which provides a fine-grained taxonomy of attack strategies grounded in LLM capabilities as well as several mitigating strategies, an overview of attack success evaluation strategies, and a framework that unifies attack-search strategies for automated red teaming. While insightful, the latter two articles' extensive lengths would be prohibitive for readers who seek a more concise overview of major red teaming concepts and trends. We see these papers as valuable in their own right, but anticipate the need for a resource that balances broad representation of the literature with concise exposition.
\section{Policies on LLM Safety}
Policies and risk mitigation strategies devised for ensuring the proper use of LLM-driven products have been crucial to their safety and success. This section serves to motivate LLM red teaming and LLM safety, providing a brief survey of some major safety considerations and policies from different leading LLM providers within industry along with risk mitigation strategies they have taken. These policies play a key role in shaping the goals for an adequate LLM safety solution, of which red teaming constitutes a critical part. 
While governments have taken steps to address LLM risks,\footnote{To date, US places no federal regulations on AI, instead leaving the matter to individual states \cite{ai_states}. Other international organizations such as the United Nations have shared some legal guidelines \cite{ai_un}. Europe recently published the AI Act, which addresses the risks of AI \cite{eu_ai}.} various organizations have established their own safety guidelines, leading to diverse priorities and approaches. As previously mentioned, organizations such as MITRE, MLCommons, and OWASP \cite{mitre_atlas, mlcommons_ai_safety_benchmarks, mlcommons_aisafety_v0_5, OWASPTop10, vidgen2024introducing} have published materials to help practitioners to secure their LLM applications; these materials form a helpful basis for policy formulation, as they categorize risks by severity and provide recommendations for evaluating AI safety . 
OpenAI was an early pioneer of LLM use policies, emphasizing legal compliance and protection of privacy \cite{openai_gpt_4_system_card, openai_safety_update, openai_ai_governance, openai_approach_to_ai_safety}, and along these ends, has employed both red teaming and guardrailing to prevent users from soliciting various kinds of harmful responses from their models \cite{openai_red_team_network}. Meta, as a large-scale social media platform, addresses risks such as election interference in their use policies \cite{meta_llama_3_responsibility, llama_responsible_use_guide, meta_sept_responsible_use_guide, meta_ai_safety_policies_safety_summit}. Anthropic's policies emphasize ethical alignment in particular \cite{anthropic_discrimination_eval, anthropic_red_teaming_methods, anthropic_hh_rlhf, anthropic_red_teaming_challenges, anthropic_frontier_threats}, and they employ guardrailing and red teaming practices and fairness evaluations to develop models with unbiased decision-making capacities.

\section{Categorizing attacks against LLMs} 
\label{sec:categorizing_attacks}

In this section, we categorize and describe various strategies for attacking LLMs. Our analysis reflects the reality that the LLM's attack surface is high context-dependent and influenced by many factors including target-system type, its infrastructure, conversational history, and access privileges. 

\subsection{Attack Methods}
Here, we categorize and describe various methods that users have employed to attack LLMs. We include a more extensive survey in the Appendix. 





\textbf{Prompt-based} attacks exploit LLMs by crafting malicious prompts to circumvent the model's safeguards. They are especially common in \textit{closed-box} systems, such as OpenAI's ChatGPT \cite{openai_gpt3.5_2023} and Google's Gemini \cite{geminiteam2024geminifamilyhighlycapable}, where attackers interact solely with the external interface of the model, lacking access to its internal weights or system-level configurations. Techniques include prompt injection \cite{liu2023prompt, mehrotra2023tree}, which disguises malicious instructions as benign inputs, and jailbreaking \cite{wei2024jailbroken, chao2024jailbreakbench}, which provoke the target LLM to ignore its safeguards. Recently, these major categories have been subdivided into a growing set of more granular categories such as indirect prompt injection \cite{greshake2023not}, refusal suppression and style injection \cite{zhou2024don, geiping2024coercing, guo2024cold}, prompt-level obfuscation \cite{pape2024prompt}, and many-shot jailbreaking \cite{anil2024many}. Some of these attacks utilize personification techniques such as role-playing to influence the target LLM into adopting a specific persona \cite{zhang-etal-2024-psysafe, shah2023scalabletransferableblackboxjailbreaks}. This manipulation can lead the LLM to relax its ethical constraints and safeguards -- e.g., \citet{chao2024jailbreakingblackboxlarge} highlight the relative effectiveness of role-playing for jailbreaking LLMs in their PAIR paradigm. Similarly, \citet{shen2024anything} introduce a notable role-playing character, DAN (Do Anything Now), which exploits the LLM's internal permissions, granting elevated privileges (e.g., Admin Privileges) to bypass safety mechanisms.

\textbf{Token-based} attacks are designed to generate variants of existing malicious prompts in order to identify novel successful attacks. Early approaches replace characters, tokens, or entire words within prompts with synonyms or symbols with comparable usage \cite{rocamorarevisiting, morris2020textattackframeworkadversarialattacks}; others simply affix symbolic material to prompts, which can confuse the system and cause it to let its guard down \cite{wallace2021universaladversarialtriggersattacking}. More recent approaches change the text encoding \cite{bai2024special}, translate it into low-resource languages \cite{wang2024languagesmattermultilingualsafety, deng2024multilingual, yong2024lowresourcelanguagesjailbreakgpt4} or use ciphers \cite{inie2024summondemonbindit, yuan2024gpt4smartsafestealthy}. By design, these strategies are not always interpretable, making it challenging to analyze how or why a specific sequence successfully bypasses the model's safeguards.  

\textbf{Gradient-based} attacks are designed instead for when attackers have access to a model's parameters -- as in an \textit{open-box} system -- such as its weights, activations, and hyperparameters. Such attacks apply gradient descent to find the most effective attack prompts \cite{shin-etal-2020-autoprompt, geisler2024attacking, wichers2024gradient}. 
A few gradient-based approaches have also shown promising generalization power when applied to closed-box systems \cite{zou2023universal}. These attacks are entirely uninterpretable and lack any semantic meaning \cite{morris-etal-2020-reevaluating}, and are commonly blocked using perplexity-based solutions \cite{jain2023baselinedefensesadversarialattacks}.

\textbf{Infrastructure} attacks involve injecting material into, extracting material from, or somehow modifying the structures that support the target LLM. One subset of such attacks includes \textit{data poisoning attacks} (or \textit{backdoor attacks}), which involve injecting problematic data or documents into the ecosystem \cite{yao2024survey, yao2024poisonprompt}. For example, the attacker might add malicious documents to an external knowledge source or API that an LLM is querying to formulate responses at runtime. The problem is often discussed in the context of agents and Retrieval Augmented Generation (RAG) pipelines — LLMs that are integrated with and call upon knowledge bases, APIs, and other software tools in order to execute tasks — since they are often susceptible to indirect prompt injection \cite{greshake2023not}, in which the malicious signal is injected into system-supporting knowledge bases or other infrastructure that then manifest an attack at retrieval time. For example, attackers could inject malicious material into external knowledge bases (e.g., Wikipedia or Wikidata) that the target LLM would then call upon to address questions. Alternatively, attackers could inject data into the model's training set, provided it is available, leading to problematic post-training behaviors. \textit{Data extraction attacks} and \textit{model extraction attacks}, on the other hand, involve extracting model data or aspects of the model itself. Data extraction attacks take place when internal data that supports the model, which may contain private or sensitive information, is unlawfully extracted \cite{carlini2021extractingtrainingdatalarge}. Beyond data, LLMs could fall prey to model theft attacks in which the model parameters themselves are extracted for unauthorized copying or use, violating intellectual property rights \cite{Kariyappa_2021_CVPR, yao2024survey}.

\subsection{Attacks by Turn Count} 
\label{sec:attack_type}


When attacking a model, we can distinguish interactions between the attacker and target LLM based on whether the attack takes place across a single turn or multiple turns. 

\textbf{Single-turn} attack pipelines are simple to implement and ideal for applications that lack memory and do not leverage conversational history \cite{xu2024redagentredteaminglarge, rawat2024attack}. A red teaming pipeline will often leverage a corpus of malicious prompts that constitute single-turn attacks (see Section \ref{sec:resources} for useful pointers to public data sources), sending pitting each them of them against the target LLM. Single-turn attacking will be limitedly effective against more complex target LLMs that critically leverage conversational history, since the latter could fall victim to attacks that only manifest after multiple conversational turns. Though common, single-turn attacks have generally become less effective now that a number of alignment techniques have been devised to ensure that the target LLM does not deviate from its intended purpose \cite{zhou2024speakturnsafetyvulnerability}.

\textbf{Multi-turn} attacks, in contrast, leverage multiple conversational turns to implement attacking. We first describe what we call the \textit{iterative attack}, which takes as a seed a single-turn attack prompt and progressively adapts it across multiple attempts at attacking, in order to maximize the likelihood of attack success. These attacks do not rely on a rich contextual history of prior interactions with the target, but instead merely track prior iterations on the same seed. Notable recent examples of iterative strategies include PAIR \cite{chao2024jailbreakbench}, TAP \cite{mehrotra2023tree}, DAN \cite{shen2024anything}, AutoDAN(-Turbo) \cite{liu2024autodangeneratingstealthyjailbreak, liu2024autodanturbolifelongagentstrategy}, RedAgent \cite{xu2024redagentredteaminglarge}, MART \cite{ge2023mart}, and APRT \cite{jiang2024automated}; early synonym-replacing approaches arguably also constitute examples \cite{morris2020textattackframeworkadversarialattacks, rocamorarevisiting}. We extend our discussion on these strategies in Section \ref{sec:execution}. Beyond the iterative attack, a multi-turn attack can be built by engaging in more complex back-and-forth conversation with the target LLM, exploiting the semantics of conversational history. To take one example from \citet{li2024llmdefensesrobustmultiturn}, in order to provoke an LLM into claiming that the health effects of Agent Orange were overstated, an attacker might: 1) ask the LLM to write an essay arguing that the substance brought about horrible health effects to victims; 2) then ask the LLM to write an essay taking the \textit{opposite} stance. The authors find that human panels are particularly effective at identifying such multi-turn attacks, well beyond the capacities of the automated approaches that they tested. 
Automated approaches that have emerged since then include Crescendo \cite{russinovich2024greatwritearticlethat}, HARM \cite{mazeika2024harmbench}, and RedQueen \cite{jiang2024red}. 

\subsection{Manual Versus Automated Attacking}
\label{sec:execution}
Attacks can be formulated manually by humans, automatically by systems such as LLMs, or by both.

\textbf{Human experts} have proven extremely helpful for red teaming LLM-driven systems \cite{li2024llmdefensesrobustmultiturn}. It has become common practice for organizations to employ human panels for red teaming and other safety-preparedness work — OpenAI, for example, employed human panels before their releases of GPT-4 \cite{openai2022moderation}. While it has been shown time and time again that humans are able to devise creative attacks, safety practitioners have found that crowdsourcing attacks can lead to templatic prompts (e.g., "give a mean prompt that begins with X") without greatly expanding attack coverage \cite{ganguli2022redteaminglanguagemodels}. Further, human annotation is expensive, which limits the number and diversity of test cases. 

\textbf{Automated solutions}, on the other hand, have gained increasing popularity by providing cheaper alternatives to evaluate the safety of LLM systems, relative to human panels. Such solutions involve automatically generating attacks against the target LLM, whose subsequent responses are evaluated for the presence of problematic content (e.g., by a trained detector). While previous work augmented attack datasets using synonym replacement and related strategies \cite{morris2020textattackframeworkadversarialattacks, rocamorarevisiting}, more recent approaches leverage LLMs to generate novel attacks \cite{perez2022red, ganguli2022redteaminglanguagemodels, deng2023attack, mo2023trustworthy, greshake2023not, yu2023gptfuzzer, paulus2024advprompterfastadaptiveadversarial, hong2024curiositydrivenredteaminglargelanguage}. In the latter case, an LLM is prompted or trained to generate a large number of examples to attack a target LLM. Since their inception, LLM-driven attack generators have been employed in whole ecosystems for automated, iterative attacking, as in PAIR \cite{chao2024jailbreakbench}, TAP \cite{mehrotra2023tree}, DAN \cite{shen2024anything}, AutoDAN(-Turbo) \cite{liu2024autodangeneratingstealthyjailbreak, liu2024autodanturbolifelongagentstrategy}, and RedAgent \cite{xu2024redagentredteaminglarge}. These solutions are unified by a common framework: an LLM-driven attacker generates an initial attack that is submitted to the target LLM; an LLM-driven evaluator then evaluates the interaction; the evaluator's signal is then passed back to the attack generator, which adapts the initial attack in some way in an attempt to increase attack success likelihood. Attack generation, response evaluation, and adaptation repeat in an iterating loop across multiple sequenced rounds. Here we single out RedAgent \cite{xu2024redagentredteaminglarge}, which additionally formulates attacks against agents that are specific to the latter's infrastructural context. Other more complex ecosystems such as MART \cite{ge2023mart} and APRT \cite{jiang2024automated} were developed based on the aforementioned iterative framework but set up an adversarial environment, in which the target LLM jointly adapts its defense strategies together with the attack generator, so that the target LLM — now possessing strengthened defenses — can be used for downstream applications. Finally, frameworks like Crescendo \cite{russinovich2024greatwritearticlethat}, HARM \cite{zhang2024holistic}, and RedQueen \cite{jiang2024red} support automated generation of more complex multi-turn attacks that exploit the semantics of the conversational history. Crescendo, for example, escalates attacks based on benign questions from prior turns — e.g., soliciting the recipe for a Molotov cocktail by first asking about its history and then about how it was historically made.


\textbf{Human-in-the-loop solutions} can involve humans guiding automated attack generation. \citet{radharapu2023aart}, for example, propose AART (AI-Assisted Red Teaming), a framework that employs automated attack generation in which humans help to select relevant attacks or filtering out those that are not likely to be successful. In addition to systems in which humans fundamentally aid AI generators, a number of AI-supported safety suites have been developed to assist \textit{humans} to efficiently conduct red teaming and identify vulnerabilities \cite{wallace2019trickcanhumanintheloopgeneration, ziegler2022adversarialtraininghighstakesreliability}.  

\section{Evaluating Attack Success}
\label{sec:eval_strategies}
The red teaming literature supplies various approaches to assessing based the target LLM's response whether an attack was successful.



\textbf{Keyword-based} (or lexical) evaluation methods attempt to match an LLM's response against a list of words, phrases, or other kind of regular expression \cite{derczynski2024garak}.
This approach is easily controllable and practitioners can expand or contract keyword lists as they see fit. On the other hand, this solution lacks insight into the general semantics of the response, and does not generalize to concepts that are not expressed in the keyword list \cite{moser2007limits}.

\textbf{Encoder-based} text classifiers provide a more robust and specializable alternative to keyword-based approaches. For example, many practitioners have trained some variety of BERT classifier \cite{devlin2019bertpretrainingdeepbidirectional, liu2019robertarobustlyoptimizedbert, caselli2021hatebertretrainingbertabusive} to detect harmful responses (e.g., \citet{yu2023gptfuzzer, derczynski2024garak}). However, these models often require training on domain-specific data or a certain kind of harm to improve performance \cite{perez2022red}, and struggle to generalize to new harms without diverse training sets \cite{askell2021generallanguageassistantlaboratory}. In contrast with LLMs-as-judges, this limits their applicability to scenarios where data are available and efficiency and cost are less of a concern. 

\textbf{LLMs-as-Judges}, on the other hand, are often leveraged due to their low barrier of entry and impressive performance \cite{zheng2023judgingllmasajudgemtbenchchatbot}. Such an approach would prompt an LLM, separate from the attack generator, to judge target system responses or even attack-response pairs (e.g., \citet{munoz2024pyrit}). 
Prior judges have returned binary assessments, scores on a 5-point scale, or continuous values \cite{shah2023scalabletransferableblackboxjailbreaks, zheng2023judgingllmasajudgemtbenchchatbot, jones-etal-2024-multi, wang-etal-2023-chatgpt}. Prompting the LLM to respond with only a quantitative judgment has been shown to limit reasoning \cite{hao2024llmreasonersnewevaluation}, and so they are often instructed to provide additional rationale \cite{sun2023safetyassessmentchineselarge, wang2023adversarial}.  Generic LLMs can perform poorly at providing domain-specific judgments (e.g., those about a financial context) \cite{dubey2024llama, jiang2024mixtralexperts} and so may require fine-tuning using extensive, annotated datasets to align the model with human intuitions \cite{rafailov2024directpreferenceoptimizationlanguage, ktomodelalignmentprospect}. LLMs also have long inference times, thus limiting adoption.

\textbf{Human reviewers} excel at providing reliable and accurate judgments due to their ability to identify subtle implications and adapt to ambiguous scenarios or domain-specific contexts \cite{ganguli2022redteaminglanguagemodels, casper2023exploreestablishexploitred}. This makes them invaluable for evaluating tasks that require subjective understanding, such as assessing content appropriateness, tone, or cultural or domain-specific subtleties. 
However, this approach faces scalability challenges as it is time-intensive, resource-demanding, and prone to bottlenecks when handling large datasets or complex tasks; Hhman evaluation can introduce variability due to personal biases, fatigue, or differences in expertise, and it is common for panelists to disagree on what constitutes a successful attack \cite{perez2022red}.

We provide in Table \ref{table:summary_table} below a summary of the aforementioned papers based their key attributes.

\begin{table*}[ht!]
    \tiny
    \resizebox{0.9\textwidth}{!}{%
    \centering
    \renewcommand{\arraystretch}{1.2}  
    \setlength{\tabcolsep}{4pt}   
    \begin{tabular}{|c|c|c|c|}
        \hline
        \rowcolor[rgb]{0.98, 0.91, 0.71}
        \textbf{Attack Method} & \textbf{Turn Count} & \textbf{Evaluation Strategy} & \textbf{Approaches} \\
        \hline
        \underline{\textbf{Prompt-based}} & \cellcolor[rgb]{1.0, 0.89, 0.88}  & \textcolor{brown}{Human Reviewers} & \cite{radharapu2023aart} \\ \cline{3-4}
          & \cellcolor[rgb]{1.0, 0.89, 0.88} & \textcolor{orange}{Keyword-based} & \cite{zhou2024don} \\ \cline{3-4}
          Prompt Injection & \multirow{-3}{*}{\cellcolor[rgb]{1.0, 0.89, 0.88} Single-turn} & LLM-as-a-Judge & \makecell{\cite{deng2023attack}, \\ \cite{shah2023scalabletransferableblackboxjailbreaks}, \cite{anil2024many}} \\ \cline{3-4}
        \cline{2-4}
         Jailbreak &  \cellcolor[rgb]{1.0, 0.98, 0.94} & \textcolor{brown}{Human Reviewers} & \cite{mehrotra2023tree}, \cite{pape2024prompt} \\ \cline{3-4}
         Style Injection &  \cellcolor[rgb]{1.0, 0.98, 0.94} & \textcolor{purple}{Encoder-based} & \makecell{\cite{yu2023gptfuzzer}, \\ \cite{hong2024curiositydrivenredteaminglargelanguage}, \cite{pape2024prompt}}\\ \cline{3-4}
          Prompt Obfuscation &  \cellcolor[rgb]{1.0, 0.98, 0.94} & \textcolor{orange}{Keyword-based} & \makecell{\cite{liu2023prompt}, \\ \cite{guo2024cold}, \cite{pape2024prompt}} \\ \cline{3-4}
         Role-playing & {\cellcolor[rgb]{1.0, 0.98, 0.94} \multirow{-4}{*}{Iterative}} & LLM-as-a-Judge & \makecell{\cite{mehrotra2023tree}, \cite{paulus2024advprompterfastadaptiveadversarial}, \\ \cite{chao2024jailbreakingblackboxlarge}, \cite{shen2024anything}, \\ \cite{liu2024autodangeneratingstealthyjailbreak}}\\ \cline{3-4}
        \cline{2-4}
        & \cellcolor[rgb]{1.0, 0.89, 0.77} & \textcolor{brown}{Human Reviewers} & \cite{ge2023mart} \\ \cline{3-4}
        & {\cellcolor[rgb]{1.0, 0.89, 0.77} \multirow{-2}{*}{Multi-turn}} & LLM-as-a-Judge & \makecell{
        \cite{russinovich2024greatwritearticlethat}, \\ \cite{ge2023mart}, 
        \cite{zhang-etal-2024-psysafe}, \\ \cite{jiang2024automated},
        \cite{zeng2024johnnypersuadellmsjailbreak},\\ \cite{jiang2024red},
        \cite{zhou2024speakturnsafetyvulnerability} \\
        } \\ \hline
        \underline{\textbf{Token-based}} & \cellcolor[rgb]{1.0, 0.89, 0.88} & \textcolor{brown}{Human Reviewers} & \makecell{\cite{yuan2024gpt4smartsafestealthy}, \cite{yong2024lowresourcelanguagesjailbreakgpt4} \\ \cite{wallace2021universaladversarialtriggersattacking}} \\ \cline{3-4}
         Encoders/Ciphers & \multirow{-2}{*}{\cellcolor[rgb]{1.0, 0.89, 0.88} Single-turn} & LLM-as-a-Judge & \cite{bai2024special}, \cite{yuan2024gpt4smartsafestealthy} \\ \cline{2-4}
         \makecell{Language Translation \\ Affix Injection} & \cellcolor[rgb]{1.0, 0.98, 0.94} Iterative  & \textcolor{purple}{Encoder-based} & \cite{rocamorarevisiting} \\ \hline
        \multirow{3}{*}{\underline{\textbf{Gradient-based}}} & {\cellcolor[rgb]{1.0, 0.89, 0.88} Single-turn} & \textcolor{orange}{Keyword-based} & \cite{zou2023universal} \\ \cline{2-4}
         & \cellcolor[rgb]{1.0, 0.98, 0.94} & \textcolor{purple}{Encoder-based} & \cite{shin-etal-2020-autoprompt}, \cite{wichers2024gradient} \\ \cline{3-4}
         & {\cellcolor[rgb]{1.0, 0.98, 0.94} \multirow{-2}{*}{Iterative}} & \textcolor{orange}{Keyword-based} & \cite{geisler2024attacking} \\ \hline
         \underline{\textbf{Infrastructure}} & \cellcolor[rgb]{1.0, 0.89, 0.88} & \textcolor{brown}{Human Reviewers} & \makecell{\cite{carlini2021extractingtrainingdatalarge}, \\ \cite{Kariyappa_2021_CVPR}}\\
          \cline{3-4}
         \makecell{Data/Model Poisoning \\ Data/Model Extraction } & \multirow{-2}{*}{{\cellcolor[rgb]{1.0, 0.89, 0.88}Single-turn}} & \textcolor{purple}{Encoder-based} & 
         \makecell{\cite{shafran2024machine}, \cite{li2024generating}, \\
         \cite{deng2024pandora},  \cite{chaudhari2024phantom}, \\
         \cite{wang2024poisoned}, \cite{pasquini2024neural}
         } \\ \cline{2-4}
        & \cellcolor[rgb]{1.0, 0.89, 0.77}Multi-turn & \textcolor{purple}{Encoder-based} & \cite{cohen2024unleashing} \\
         \hline
         
    \end{tabular}
    }
    \centering
    \caption{Overview of red teaming papers categorized by key attributes.}
    \label{table:summary_table}
\end{table*}

\section{Safety Metrics}
\label{sec:evaluation_metrics}
There exist various ways to measure overall model safety in the context of a red teaming experiment. \footnote{These metrics do not address \textit{hallucinations} i.e., incorrect or misleading results that LLMs may generate. However, there are still scenarios where hallucination may cause harm without a malicious intention.} 

\textbf{Attack Success Rate} (ASR) is a popular metric employed to gauge the effectiveness of a red teaming strategy, defined as the ratio of successful attacks to total attempts \cite{zou2023universal, russinovich2024greatwritearticlethat, shen2024anything}. ASR has conventionally indexed on a narrow notion of safety, failing to consider the relevance or usefulness of target responses as they pertains to a specific context. To address this limitation, \citet{jiang2024automated} introduced a new metric, \textit{Attack Effectiveness Rate} (AER), that evaluates collective responses along both safety and response helpfulness. Other substantive metrics have arisen to capture the different dimensions of safety. \textit{Toxicity} (or {Harmfulness) is computed by evaluating whether the generated responses contain specific harmful content like killing a person or robbing a bank \cite{xu2023cvaluesmeasuringvalueschinese, zeng2024johnnypersuadellmsjailbreak}. \textit{Compliance} (or Obedience) measures compliance of a model to the instructions in a malicious prompt \cite{jin2024attackeval, yu2023gptfuzzer}. For example, in \cite{yu2023gptfuzzer}, the authors assess responses along a 4-point compliance scale ranging from full refusal to full compliance. \textit{Relevance} refers to the pertinence of the model's response to the attack prompt. If a model output contains generic details, but fails to be relevant, then it should be termed as an unsuccessful attack. Practitioners have employed humans or even LLMs (e.g., \citet{Takemoto_2024}) to assess the relevance of a response relative to an input. Fluency, calculated using measures of model perplexity, is often assessed jointly with relevance for a more comprehensive assessment the target system's response \cite{khalatbari2023learnlearngenerativesafety}. Any of the aforementioned substantive metrics can be assessed manually or automatically (e.g., by an LLM-as-a-judge).



\section{Public Red Teaming Resources}
\label{sec:resources}
Several datasets and libraries have been developed to facilitate the quick development of LLM red teaming applications by the research community. 

\textbf{Frameworks} like Pyrit \cite{munoz2024pyrit}, for example, pit an attacker system against a target, with attack-response pairs judged by an evaluator. Pyrit is designed with a low barrier to entry and enables easy integration of new attack strategies. Garak \cite{derczynski2024garak} provides a similar framework, and offers advanced logging and report generation capabilities. Giskard \cite{giskard}, an enterprise level framework, offers scalability. Multi-round Automatic red teaming (MART) \cite{ge2023mart} as described in Section \ref{sec:attack_type} represents another state-of-the-art adversarial multi-turn framework.

\textbf{Datasets} have also been curated by the research community for probing LLM vulnerabilities to support red teaming efforts. These resources are often paired with a research paper describing their creation process. One such dataset is JailbreakBench \cite{chao2024jailbreakbench}, which focuses on prompts designed to elicit behaviors that violate OpenAI's usage policies, covering areas like harassment, malware, and disinformation. Another dataset, GPTFuzzer \cite{yu2023gptfuzzer}, includes prompts and questions aimed at identifying vulnerabilities in LLMs, with a focus on generating harmful or unsafe responses.
ALERT \cite{tedeschi2024alert} offers a comprehensive benchmark for assessing LLM safety through red teaming, with a collection of instructions and questions categorized by the level of harm involved. SafetyBench \cite{zhang2023safetybench} includes multiple-choice questions designed to test knowledge on safety and identify potential risks.
XSafety \cite{wang2024languagesmattermultilingualsafety} covers commonly used safety issues across multiple languages, providing a valuable resource for evaluating multilingual LLMs.
\citep{shen2024anything} also released DAN, a popular dataset for evaluating in-the-wild jailbreak prompts that includes prompts targeting behaviors disallowed by OpenAI -- the attacks in this dataset have been sourced online from public forums. DoNotAnswer \cite{wang-etal-2024-answer} evaluates ``dangerous capabilities'' of LLMs by assessing their responses to questions that should ideally not be answered. HarmBench \cite{mazeika2024harmbench} evaluates the effectiveness of automated red teaming methods with a focus on different semantic categories of harmful behavior. \citet{li2024llmdefensesrobustmultiturn} supply Multi-Turn Human Jailbreaks (MHJ), a dataset of human-formulated multi-turn jailbreaks. Finally, DecodingTrust \cite{wang2023decodingtrust} evaluates the trustworthiness of LLMs across various perspectives, including toxicity, stereotypes, and privacy. 
Several other resources are listed by other organizations such as the UK AI Safety Institute \cite{ukai}.



\section{Mitigation Strategies}

While red teaming probes systems for vulnerabilities, guardrailing safeguards an application after its deployment. Here we present a few approaches to integrating guardrails into the LLM system.


\textbf{System prompts} are carefully crafted to guide the LLM away from engaging with unsafe inputs and returning harmful responses (e.g., \citet{openai2024gpt4technicalreport, jiang2023mistral7b}). \citet{zheng2024on} suggest that LLMs refuse to respond to inputs more readily when they are supplied a safety prompt, even when the input is harmless. Other approaches automate generation of safety prompts -- e.g., \citet{zou2024messagereallyimportantjailbreaks} propose a genetic algorithm for generating safety prompts that best protect against jailbreaks.


\textbf{Content Filtering} approaches delegate safeguarding to other systems that serve to filter model inputs and/or outputs. For example, PromptGuard \cite{grattafiori2024llama3herdmodels} is a BERT-based classifier fine-tuned on a large corpus of prompt injections and jailbreaks. \citet{jain2023baselinedefensesadversarialattacks} present perplexity filtering, which detects incoherence, as an effective defense against token-based attacks, and also propose a paraphrasing technique that rephrases adversarial inputs in such a way that the safe instructions are preserved but adversarial tokens are reproduced inaccurately. Llama Guard \cite{inan2023llama} is a fine-tuned LLM that classifies for potential risks in user prompts and model responses based on their safety policies. AutoDefense \cite{zeng2024autodefensemultiagentllmdefense} is a multi-agent framework that leverages multiple LLM agents to collaboratively protect against attacks. 
OpenAI also provides a proprietary API \cite{openai_moderation_api} that can be used to classify content according to its defined moderation taxonomy. 
These approaches are promising for single-turn attacks, but may be vulnerable to multi-turn attacks that conceal malicious intent across multiple turns to avoid detection.

\textbf{Fine-tuning and alignment} can enhance the safety alignment of LLMs. Supervised Fine-Tuning (SFT) can be applied with high-quality safety data (pairs of harmful instructions/attacks and refusal responses) in order to improve model robustness \cite{touvron2023llama2openfoundation}. Reinforcement Learning from Human Feedback (RLHF) is useful for further safety alignment, and has minimal performance impact \cite{ouyang2022traininglanguagemodelsfollow}. It first fits a reward model that captures human preference, using it for reinforcement learning to teach the target model to maximize this estimated reward. Variations of vanilla RLHF, such as Direct Preference Optimization (DPO) \cite{rafailov2024directpreferenceoptimizationlanguage, rad2025refininginputguardrailsenhancing} and Distributional Preference Learning (DPL) \cite{siththaranjan2024distributionalpreferencelearningunderstanding} have also demonstrated reductions in jailbreak risks.
Fine-tuning an LLM also makes it immune to gradient-based attacks which rely on the knowledge of the model's internal weights. 


\section{Conclusion and Future Directions}
This paper provided a survey of the fast-evolving, multifaceted arena of LLM red teaming. 
We first described some of the major safety-related considerations that large tech companies faced as they were building out their LLMs.
We then provided a synopsis of the conventional red teaming pipeline, a deep dive into its key components and supporting methodologies for attacking, evaluating attack success, and safety metrics for measuring experimental outcomes. We shared public resources that practitioners can leverage to develop their own pipelines. Finally, we outlined  popular guardrailing strategies that can be put in place to protect applications from unexpected attacks.

In the future, we anticipate more research on automated multi-turn red teaming, addressing \citet{li2024llmdefensesrobustmultiturn}'s observation that humans vastly outperform automated solutions in this area presently. In addition, we look forward to more research on adapting automated attackers to generate sets of attacks that are both diverse and relevant to a given target system; such approaches might involve fine-tuning \cite{hong2024curiositydrivenredteaminglargelanguage, lee2024learningdiverseattackslarge}, a separate strategizing model \cite{liu2024autodanturbolifelongagentstrategy}, a sophisticated search algorithm \cite{chao2024jailbreakbench}, or something entirely new — e.g., adapting generation by identifying which prompts tend to bring about the best attacks once served to the generator. We also look forward to advances in frameworks in which multiple LLMs interact or compete, as in PAIR or MART \cite{ge2023mart, chao2024jailbreakbench}; we see these systems as paving the way toward continuous monitoring and adaptive security.
Finally, we anticipate that establishing a diverse array of standardized metrics will be critical for comparing approaches and measuring progress. 
\section{Limitations}
This paper provides a concise overview on the current red teaming literature. However, we acknowledge that due to space limitations  -- we prioritized mentioning the most impactful and cited papers in the field -- the paper could miss mentioning some relevant works.
We would like to highlight how red teaming alone does not guarantee 
the safety of a model after deployment. There may be outside factors or new research breakthroughs that could impact the safety of models after they have been deployed and we therefore recommend a constant monitoring of such systems in production. Additionally, to ensure the safety of an LLM system, we underscore again the importance of guardrailing solutions that constitute an additional line of defense against malicious actors. 
Finally, as the regulation space and technology use evolve, we cannot exclude the emergence of additional risks associated to LLM usage that we did not anticipate at the time of writing.

\bibliography{main}

\section*{Appendix}

\definecolor{floralwhite}{rgb}{1.0, 0.98, 0.94}

\section{Attack Methods}
\label{sec:attack_methods_detailed}
\subsection{Gradient-based}

\textbf{GCG}. In Greedy Coordinate Gradient (GCG) \cite{zou2023universal}, token-level optimization is applied to an adversarial suffix, appended to a user prompt to create a test case. This suffix is fine-tuned to maximize the log probability assigned by the target LLM to an affirmative target string, which triggers the desired behavior.

\textbf{PGD}. In the Projected Gradient Descent (PGD) \cite{geisler2024attacking} paper, the authors demonstrate that PGD for LLMs achieves effectiveness comparable to discrete optimization methods while significantly improving efficiency. They introduce a novel approach that continuously relaxes the process of adding or removing tokens, enabling optimization over variable-length sequences. Furthermore, the paper is the first to highlight and analyze the trade-off between cost and effectiveness in the context of automatic red teaming, providing valuable insights into optimizing adversarial techniques for language models. They claim to show performance boost over GCG.

\textbf{AutoPROMPT}. AutoPROMPT \cite{shin-etal-2020-autoprompt} employs an automated method to create attack prompts for a set of tasks based on a gradient-guided search on Masked Language Models (MLMs) like Roberta \cite{liu2019robertarobustlyoptimizedbert}. AutoPROMPT generates prompts by combining the original task inputs with a predefined set of trigger tokens structured according to a template. These tokens are optimized using a variant of the gradient-based search strategy.

\section{Attack Types
\label{sec:attack_type_detailed}
}
Here we provide more details and information about a few attack types that are most commonly used in the red teaming literature.

\subsection{Iterative}

\textbf{PAIR}. Prompt Automatic Iterative Refinement (PAIR) \cite{chao2024jailbreakingblackboxlarge} employs a separate attacker language model to generate jailbreaks for any target model. The attacker model is provided with a detailed system prompt instructing it to act as a red teaming assistant. Using in-context learning, PAIR iteratively refines candidate prompts by incorporating prior attempts and the target model's responses into the chat history until a successful jailbreak is achieved. Additionally, the attacker model reflects on both the previous prompt and the target model's response to produce an "improved" prompt, leveraging chain-of-thought reasoning. This approach enhances model interpretability by enabling the attacker model to explain its reasoning and strategies.

\textbf{TAP}. Tree of Attack with Pruning (TAP) \cite{mehrotra2023tree} utilizes three LLMs: an attacker tasked with generating jailbreaking prompts using tree-of-thoughts reasoning, an evaluator responsible for assessing these prompts and determining the success of the jailbreak attempt, and a target, which is the LLM being subjected to the jailbreak attempt. TAP is a generalization of the PAIR method:  TAP specializes to PAIR when its branching factor is 1 and pruning of off-topic prompts is disabled.

\textbf{AutoDAN}. AutoDAN \cite{liu2024autodangeneratingstealthyjailbreak} generates jailbreak prompts using a hierarchical genetic algorithm. From an initial population of attack prompts, sentence- and paragraph-level crossovers, along with LLM-powered rephrasing, are applied to produce subsequent generations of attacks. The fitness function measures the probability of affirmative response tokens, the same as \cite{zou2023universal}. Fluency of the resulting attacks is preserved, which means that perplexity-based mitigation methods are generally ineffective.

\subsection{Multi-Turn}
\textbf{Crescendo}. Crescendo \cite{russinovich2024greatwritearticlethat} leverages an LLM's intrinsic ability to identify patterns and emphasize recent context, particularly the text generated within the conversation. The approach begins with an innocuous abstract query related to the targeted jailbreaking objective. Over successive interactions, Crescendo incrementally steers the model toward producing harmful outputs through small, seemingly benign steps. However, as Crescendo relies heavily on maintaining historical context to construct its attacks, models that do not retain conversational history or have limited context windows are inherently more resistant to this technique.

\textbf{HARM}. HARM \cite{zhang2024holistic} employs a top-down methodology, relying on a detailed and defined risk taxonomy to generate various test cases. It incorporates a fine-tuning strategy and reinforcement learning (from manual red teaming and human feedback) to facilitate multi-turn adversarial probing. 

\textbf{PAP}. Persuasive Adversarial Prompts (PAP) \cite{zeng2024johnnypersuadellmsjailbreak} develops a persuasion taxonomy and employs persuasion technique to jailbreak where an attacker LLM tries to make the request sound more convincing according to persuasive strategy.

\end{document}